%% file: main.tex
\documentclass[anonymous]{article}
\usepackage{ijcai25}
\usepackage{times}
\usepackage{soul}
\usepackage{url}
\usepackage[hidelinks]{hyperref}
\usepackage[utf8]{inputenc}
\usepackage[small]{caption}
\usepackage{graphicx}
\usepackage{amsmath}
\usepackage{amsthm}
\usepackage{booktabs}
\usepackage{algorithmic}
\usepackage[switch]{lineno}

\usepackage{graphicx} 
\usepackage{caption} 
\usepackage{algorithmic}

\usepackage{amsthm}
\usepackage{amsmath}
\usepackage[switch]{lineno}
\usepackage{amsfonts}   
\usepackage{multirow}
\usepackage{caption}
\usepackage{xcolor}
\usepackage{subfigure}

\newtheorem{problem}{Problem}

\usepackage{booktabs}  
\usepackage{amsthm, amsmath, amssymb, amsthm}
\usepackage{booktabs} 
\usepackage{multicol,multirow}
\usepackage{makecell}
\usepackage{amsfonts}       
\usepackage{nicefrac}       
\usepackage{microtype}      
\usepackage{xcolor}         
\usepackage{amsthm}
\usepackage{colortbl}
\usepackage[ruled,vlined]{algorithm2e}

\newcommand{\thgcn}{\textsc{THeGCN}}

\title{\thgcn: Temporal Heterophilic Graph Convolutional Network}
\author{
Yuchen Yan$^1$
\and
Yuzhong Chen$^2$\and
Huiyuan Chen$^2$\And
Xiaoting Li$^2$\And
Zhe Xu$^1$\And
Zhichen Zeng$^1$\And
Lihui Liu$^1$\And
Zhining Liu$^1$\And
Hanghang Tong$^1$
\\
\affiliations
$^1$University of Illinois at Urbana-Champaign\\
$^2$Visa Research\\
\emails
\{yucheny5,zhexu3,zhichenz,lihuil2,liu326,htong\}@illinois.edu,
\{yuzchen,hchen,xiaotili\}@visa.com,
}

\begin{document}

\maketitle

\begin{abstract}
\input{01abs.tex}
\end{abstract}

\input{02intro.tex}
\input{04prelinminary.tex}
\input{05model.tex}
\input{06experiment.tex}
\input{03related_work.tex}
\input{07conclusion.tex}


\input{08appendix.tex}
\bibliographystyle{named}
\bibliography{main}
\end{document}

%% file: 01abs.tex

Graph Neural Networks (GNNs) have exhibited remarkable efficacy in diverse graph learning tasks, particularly on static homophilic graphs. Recent attention has pivoted towards more intricate structures, encompassing (1) static heterophilic graphs encountering the \textit{edge heterophily issue} in the spatial domain and (2) event-based continuous graphs in the temporal domain. State-of-the-art (SOTA) has been concurrently addressing these two lines of work but tends to overlook the presence of heterophily in the temporal domain, constituting the \textit{temporal heterophily issue}. Furthermore, we highlight that the \textit{edge heterophily issue} and the \textit{temporal heterophily issue} often co-exist in event-based continuous graphs, giving rise to the \textit{temporal edge heterophily challenge}. To tackle this challenge, this paper first introduces the temporal edge heterophily measurement. Subsequently, we propose the \underline{T}emporal \underline{He}terophilic \underline{G}raph \underline{C}onvolutional \underline{N}etwork (\thgcn), an innovative model that incorporates the low/high-pass graph signal filtering technique to accurately capture both edge (spatial) heterophily and temporal heterophily. Specifically, the \thgcn\ model consists of two key components: a sampler and an aggregator. The sampler selects events relevant to a node at a given moment. Then, the aggregator executes message-passing, encoding temporal information, node attributes, and edge attributes into node embeddings. Extensive experiments conducted on 5 real-world datasets validate the efficacy of \thgcn.

%% file: 02intro.tex
\section{Introduction}
    


\footnote{Preliminary work using IJCAI template.} Graph Neural Network (GNN) \cite{kipf2016semi,velivckovic2017graph,hamilton2017inductive} has exhibited great power in a large number of graph learning tasks, such as node classification \cite{kipf2016semi}, link prediction \cite{zhang2018link}, recommender systems \cite{wang2022rete} and many more.
The \emph{homophily assumption}, i.e., the connected node pairs tend to share the same labels, is widely adopted by many existing GNNs. Beyond that, most GNNs are designed for static graphs, whose topology and attributes are fixed once the graph is constructed.
\begin{figure}[htbp]
    \begin{center}
    \subfigure[Edge homophily.]{\includegraphics[scale=0.42]{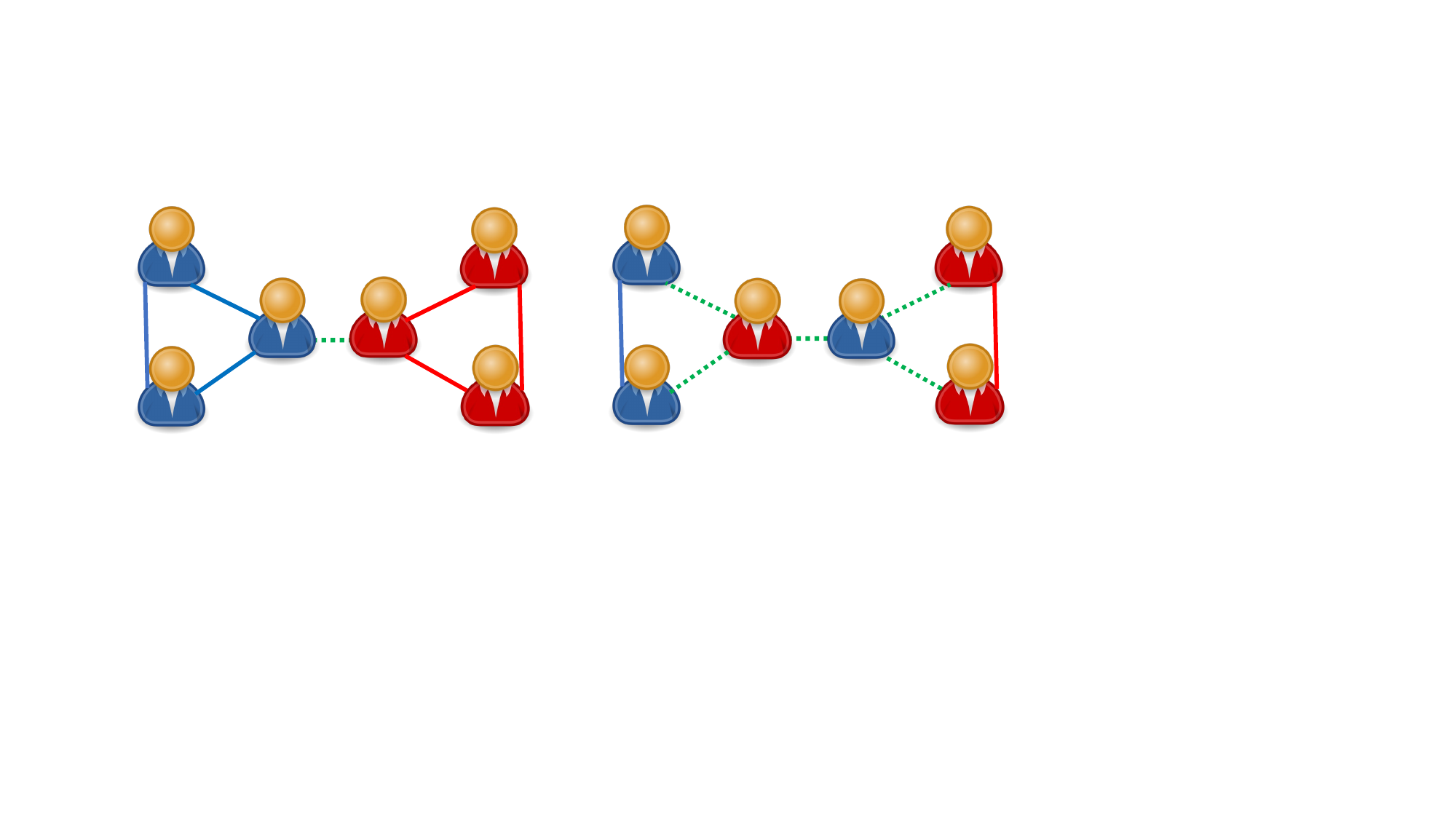}}
    \subfigure[Edge heterophily.]{\includegraphics[scale=0.42]{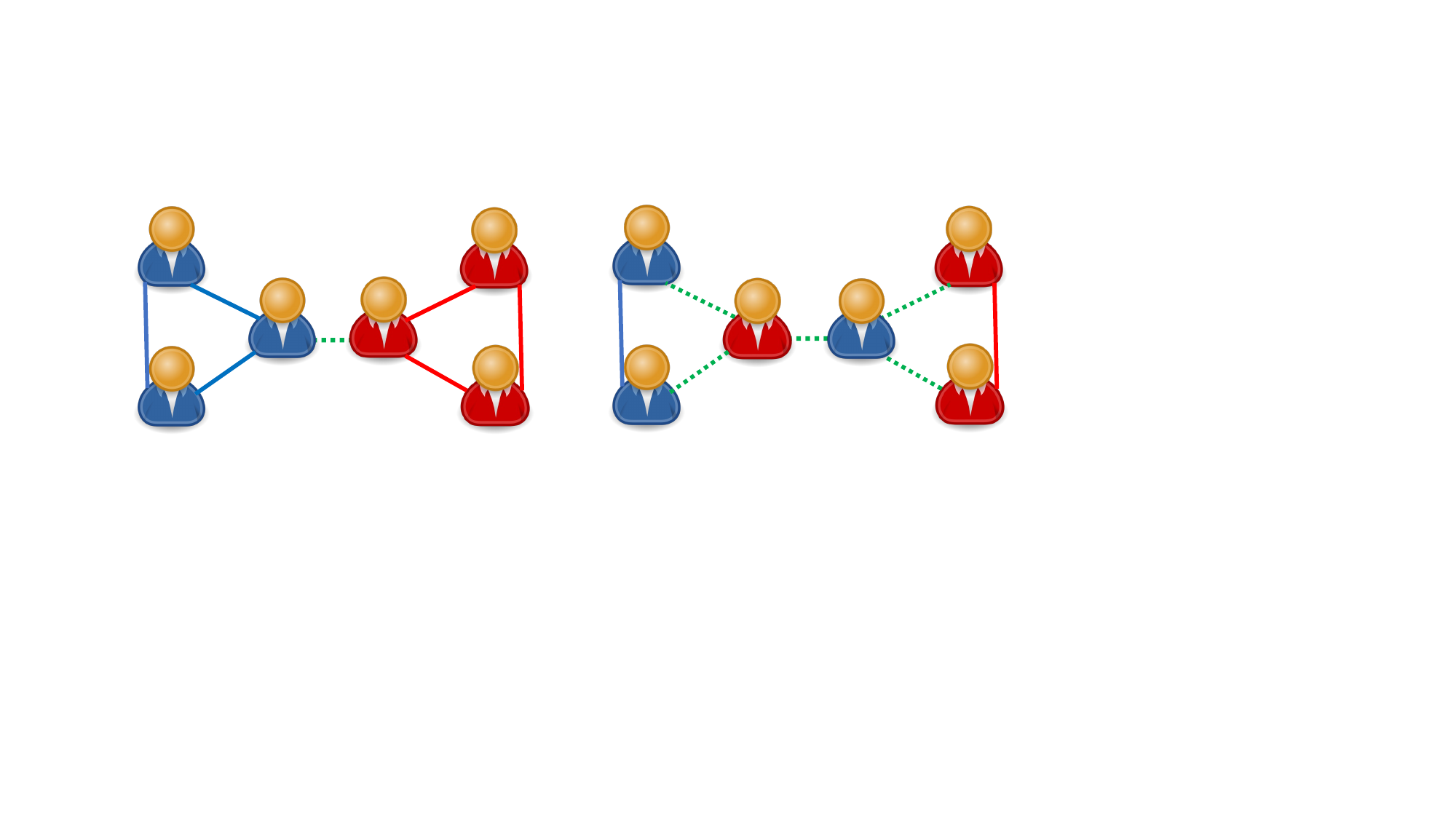}}
        \caption{Examples of edge homophily (a) and edge heterophily (b). In (a), nodes with same labels (i.e., colors) tend to connect (solid lines), while in (b), most edges (dashed lines) link nodes with different labels.}
        \label{fig:ehom}
        \end{center}
    \end{figure}
Recently, graphs with more complex topology and attributes, such as heterophilic graphs and temporal graphs, have attracted increasing research attention. In the spatial domain, it is found that mainstream GNNs have difficulties handling the \textit{edge heterophily issue}, which refers to the phenomenon that connected nodes may have distinct labels/classes in real-world graphs~\cite{bo2021beyond,luan2021heterophily}. Examples of edge homophily and edge heterophily are shown in Figure~\ref{fig:ehom}. Numerous GNNs \cite{li2022finding,du2022gbk,zheng2022graph} have been proposed to solve the \textit{edge heterophily issue}. For example, H2GCN \cite{zhu2020beyond} samples distant neighbors for message-passing. FAGCN \cite{bo2021beyond} and GPRGNN \cite{chien2020adaptive} use adaptive low/high-pass filters to make connected node pairs share similar/disparate embeddings. On the other hand, in the temporal domain, the real-world graphs keep evolving over time due to the ever-emerging events. To model such event-based continuous graphs, TGAT \cite{xu2020inductive} and TGN \cite{rossi2020temporal} transfer the homophily assumption from static GNNs to temporal graphs and utilize the attention technique for message-passing. To summarize, as shown in Table~\ref{tb:algexamples}, SOTA has been addressing these two lines of work almost in parallel:
\begin{itemize}
{\em 
    \item [1.] Existing heterophilic GNNs almost exclusively focus on static graphs;
    \item [2.] Existing temporal GNNs almost exclusively rely on the edge homophily assumption. 
}
\end{itemize} 

\begin{table}[t]
\centering
\scalebox{0.6}{
\begin{tabular}{|c|c|c|}
\toprule
Models & Static & Temporal\\
\midrule
 Homophilic & \makecell[c]{GCN \cite{kipf2016semi} \\ GAT \cite{velivckovic2017graph}\\ SGC \cite{wu2019simplifying}} & \makecell[c]{TGN \cite{rossi2020temporal}\\ TGAT \cite{xu2020inductive} \\ JODIE \cite{kumar2019predicting}\\} \\
 \hline
Heterophilic & \makecell[c]{FAGCN \cite{bo2021beyond}\\ GPRGNN \cite{chien2020adaptive}\\ HOG-GCN \cite{wang2022powerful}} & \makecell[c]{\textcolor{red}{\textbf{\thgcn}} \\ \textcolor{red}{(This paper)}}\\ 
\bottomrule
\end{tabular}}
\caption{Some representatives of (1) static homophilic GNNs; (2) temporal homophilic GNNs; and (3) static heterophilc GNNs.}
\label{tb:algexamples}
\end{table} 

\begin{figure}[t]
    \begin{center}
    \subfigure[Temporal homophily.]{\includegraphics[scale=0.26]{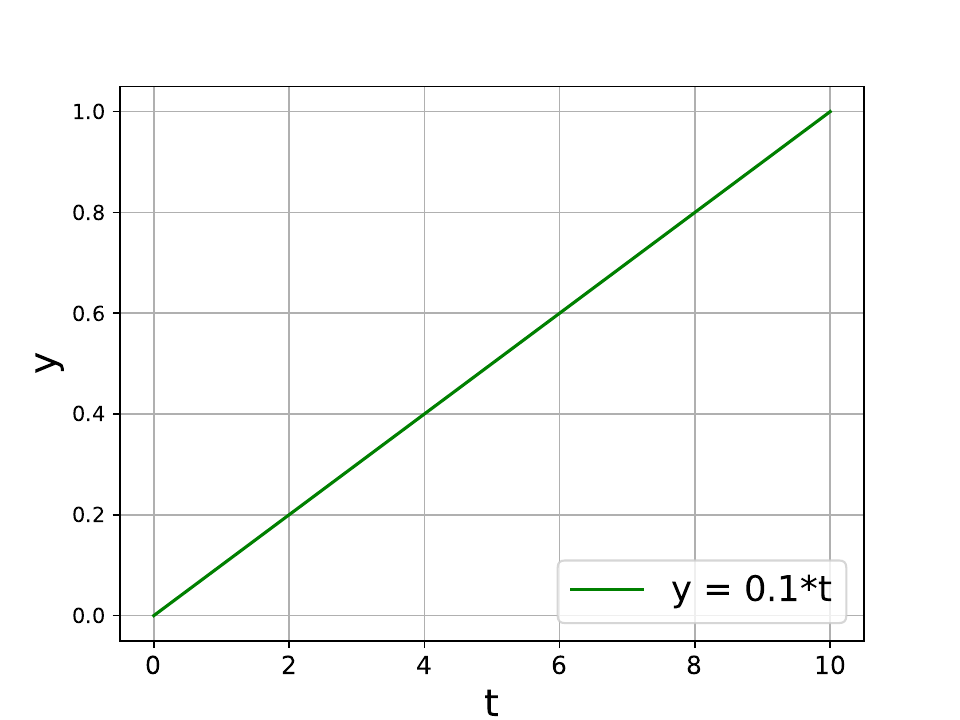}}
    \hspace{-12pt}
    \subfigure[Temporal heterophily.]{\includegraphics[scale=0.26]{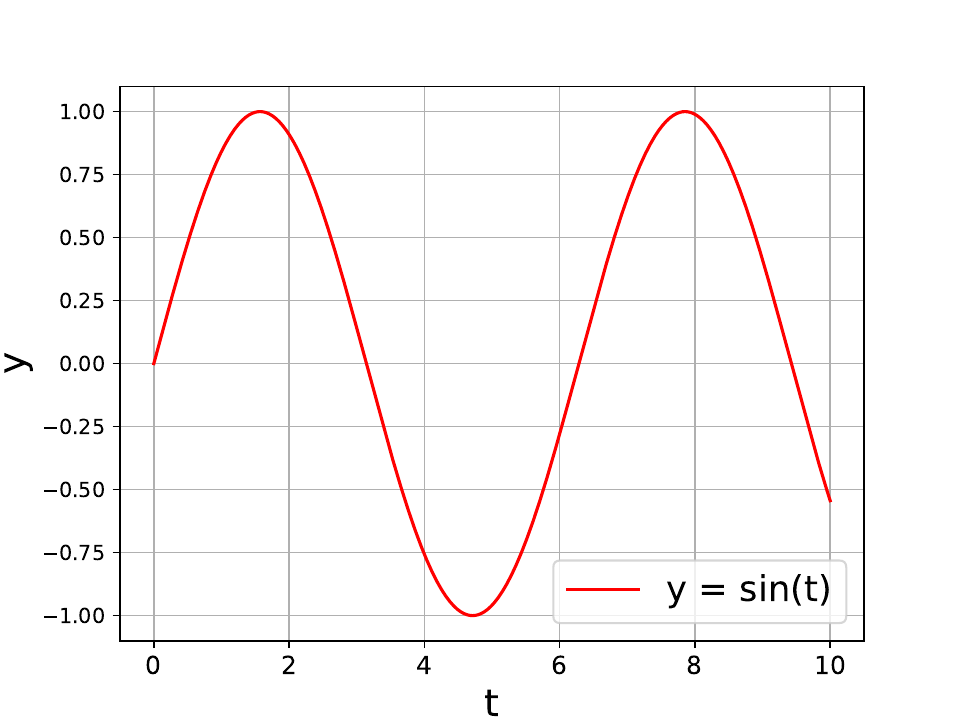}}
        \caption{Examples of temporal homophily and temporal heterophily.}
        \label{fig:tp}
        \end{center}
    \end{figure}

However, in real-world graphs, the heterophily issue could occur not only in the spatial domain but also in the temporal domain, the latter of which is referred to as the \textit{temporal heterophily issue} in this paper. Usually, it is more likely that recent events tend to own similar observations to the present event, which is precisely the \textit{temporal homophily assumption} behind most temporal graph neural networks implicitly. Nonetheless, this assumption may not always be true. As the illustrative example shown in Figure~\ref{fig:tp}, if we consider three timestamps: $t=5$, $t=8$ and $t=10$, the linear temporal pattern in Figure~\ref{fig:tp} (a) follows the temporal homophily assumption, e.g., $y(8)$ is more similar to $y(10)$ compared with $y(5)$, whereas the periodic temporal pattern in Figure~\ref{fig:tp} (b) demonstrates that $y(10)$ is more similar to $y(5)$ than $y(8)$, which is a special case of the temporal heterophily. In addition to periodic temporal patterns, spike (e.g., abnormal events) is another case of the temporal heterophily. 

\begin{figure*}[t]
    \centering
    \includegraphics[width=\linewidth]{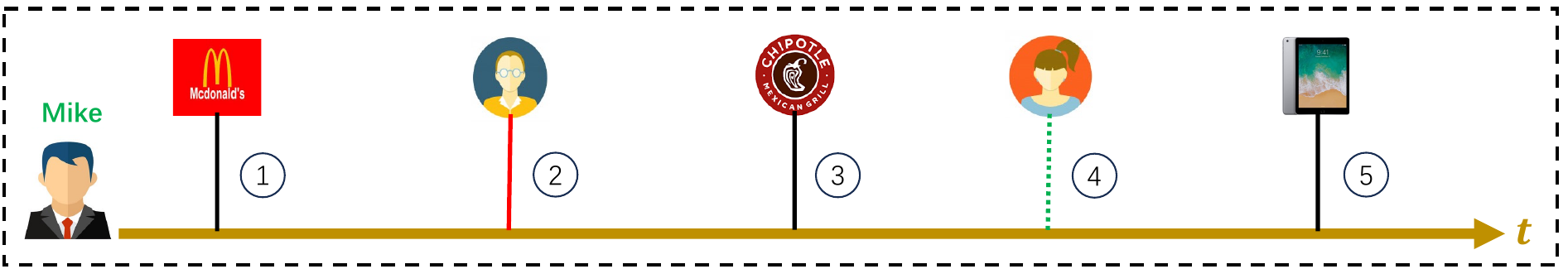}
    \caption{An event-based continuous graph: one day of Mike. Interactions in this figure: \textcircled{1}, 8:28 a.m., buy a muffin from Macdonald's; \textcircled{2}, 9:43 a.m., reply to Tom on Twitter to express agreement; \textcircled{3}, 12:15 p.m., buy a bowl from Chipotle; \textcircled{4}, 2:45 p.m., reply to Lucy on Twitter to express disagreement; \textcircled{5}, 4:45 p.m., buy an iPad from Apple.}
    \label{fig:example1}
\end{figure*}

Furthermore, the \textit{edge heterophily} (spatial) and the \textit{temporal heterophily} often co-exist in real-world applications. Take one day of Mike in Figure~\ref{fig:example1} as an example. Regarding the \textit{edge heterophily issue}, Mike replies to Tom on Twitter at 9:43 a.m. to express \textit{agreement} (i.e., edge homophily), while replies to Lucy at 2:45 p.m. to express \textit{disagreement} (i.e., edge heterophily). For the \textit{temporal heterophily issue}, Tom buys a muffin at 8:28 a.m. and buys an iPad at 4:45 p.m. If we predict the event at 7:45 a.m. next day, it is more likely that Mike will buy another muffin than a new iPad even though the 4:45 p.m. event (i.e., buying an iPad) is more recent. From this example, it is evident that the temporal heterophily and the edge heterophily are tightly coupled. The coupling makes it highly challenging, if not infeasible, for existing temporal graph neural networks to capture both the temporal heterophily and the edge heterophily accurately. We refer to the co-existence/coupling of the \textit{edge heterophily issue} and the \textit{temporal heterophily issue} as the \textit{temporal edge heterophily challenge}.
In this paper, we first define the temporal edge heterophily measurement, a generalization of existing static edge heterophily measurement. Then, we propose a simple yet highly effective model \thgcn\ to solve the \textit{temporal edge heterophily challenge} for complex event-based continuous graphs. Specifically, the key idea of \thgcn\ is adopting the low/high-pass signal filtering technique to handle the co-existence of the \textit{edge heterophily issue} and the \textit{temporal heterophily issue}. The \thgcn\ model consists of two components: a \textit{sampler} and an \textit{aggregator}. The \textit{sampler} samples events related to the target node at a specific timestamp. Based on these sampled events, we combine all these interactions related to the target node to build a temporally sampled graph. Then, the {\em aggregator} conducts message-passing on the temporally sampled graph, which encodes all relevant temporal information, node attributes, and interaction/edge attributes into the node embedding. Extensive experiments on 5 real-world datasets demonstrate the effectiveness of the proposed \thgcn\ framework.
The main contributions of this paper are summarized as follows:
\begin{itemize}
    \item \textbf{Problem formulation.} To our best knowledge, this is the first work to study the temporal edge heterophily challenge on event-based continuous graphs. 
    \item \textbf{Succinct yet effective model.} A succinct yet highly powerful model \thgcn\ is proposed to jointly handle the temporal and edge heterophily issues together.
    \item \textbf{Extensive experiments.} \thgcn\ consistently achieves superior performance on semi-supervised temporal node classification tasks on 5 real-world datasets.
\end{itemize}

%% file: 04prelinminary.tex
\section{Preliminaries and Problem Definition} \label{sec:pre}
In this section, we introduce the notations and review the preliminaries about the graph convolutional network (GCN) and edge heterophily. Based on that, the definition of semi-supervised node classification on event-based continuous graphs is introduced. 

\textbf{Notations.}
The main symbols and notations used in this paper are shown in Table~\ref{tb:notation}. We utilize bold uppercase letters for matrices (e.g., $\mathbf{A}$), bold lowercase letters for column vectors (e.g., $\mathbf{u}$) and lowercase letters for scalars (e.g., $\alpha$). We use the superscript $\top$ for the transpose of matrices and vectors (e.g., $\mathbf{A}^{\top}$ and $\mathbf{u}^{\top}$).

\textbf{Graph convolutional network (GCN).} 
Graph convolutional network (GCN) is a widely used tool for graph embedding. An attributed static undirected graph $\mathcal{G} = \{\mathbf{A}, \mathbf{X}\}$ contains an adjacency matrix $\mathbf{A}$ and a node attribute matrix $\mathbf{X}$.  $\mathbf{D}$ denotes the diagonal degree matrix of $\mathbf{A}$. The adjacency matrix with self-loops is given by $\tilde{\mathbf{A}}=\mathbf{A}+\mathbf{I}$ ($\mathbf{I}$ is the identity matrix), and all variables derived from $\tilde{\mathbf{A}}$ are decorated with symbol~ $\tilde{}$~, e.g., $\Tilde{\mathbf{D}}$ represents the diagonal degree matrix of $\tilde{\mathbf{A}}$. The parameter and node embedding matrices in the $l$-th layer of a GCN are denoted by $\mathbf{W}^{(l)}$ and $\mathbf{H}^{(l)}$, respectively.

The layer-wise message-passing and aggregation of GCN \cite{kipf2016semi} is given by
 \begin{equation}
     \mathbf{H}^{(l+1)} = \sigma(\Tilde{\mathbf{D}}^{-\frac{1}{2}}\Tilde{\mathbf{A}}\Tilde{\mathbf{D}}^{-\frac{1}{2}}\mathbf{H}^{(l)}\mathbf{W}^{(l)}),
     \label{eq:gcn}
 \end{equation}
where $\mathbf{H}^{(l)}$/$\mathbf{H}^{(l+1)}$ stands for the embedding matrix ($\mathbf{H}^{(0)}=\mathbf{X}$) in the $l$-th/$(l+1)$-th layer; $\mathbf{W}^{(l)}$ is the trainable parameter matrix; and $\sigma(\cdot)$ is the non-linear activation function. 


\begin{table}[t]
\centering
\caption{Symbols and Notations.}
\begin{tabular}{|c|c|}
\hline
\textbf{Symbol}& \textbf{Definition}\\
\hline
$\mathcal{G}$ & a static undirected graph\\
$\mathcal{T}$ & an event-based continuous graph\\
$\mathcal{N}$ & node set of $\mathcal{T}$\\
$\mathcal{E}$ & interaction set of $\mathcal{T}$\\
\hline
$\mathbf{A}$ & adjacency matrix\\
$\mathbf{X}$ & attribute matrix\\
$\mathbf{I}$ & identity matrix\\
$\mathbf{H}$ & embedding matrix\\
$\mathbf{D}$ & degree matrix\\
$\mathbf{W}$ & parameter matrix\\
\hline
$l$ & \# of layers\\
$f$ & dimensions of hidden layers\\
$v_i$ & the $i$-th node\\
$t, t'$ & time variable\\
$v_i^t$ & $v_i$ at $t$\\
$e^t_{v_i, v_j}$ & event/interaction between $v_i$ and $v_j$ at $t$\\
$y_{v_i}$ & label/class of $v_i$\\
$y_{v_i}^{t}$ & label/class of $v_i$ at $t$\\
$H(\cdot)$ & heterophily measurement\\
$E(\cdot)$ & time encoder\\
$p,q$ & low-pass/high-pass attention weight\\
$h_{\textrm{max}}$ & maximal hop for sampling\\
$N_{\textrm{max}}$ & maximal neighbors for sampling\\
\hline
$\mathbf{x}_{v_i^t}$ & attribute vector of $v_i$ at $t$\\
$\mathbf{m}_{e^t_{v_i, v_j}}$ & attribute vector of $e^t_{v_i, v_j}$\\
$\mathbf{h}_{v_i}^{(l)}$ & embedding of $v_i$ at layer $l$\\
$\mathbf{w}$ & parameter vector\\
\hline
\end{tabular}
\label{tb:notation}
\end{table}
\textbf{Edge heterophily.} Edge heterophily describes to what extent edges tend to link nodes with disparate labels. The edge heterophily measurement ~\cite{zhu2020beyond} for static graphs is defined as~\cite{zhu2020beyond}: $H(\mathcal{G})=\frac{\sum_{i,j,\mathbf{A}[i,j]=1}\langle y_{v_i} \neq y_{v_j}\rangle}{\sum_{i,j}\mathbf{A}[i,j]}\in[0,1],$
where $\langle x \rangle=1$ if $x$ is true and $0$ otherwise. A graph is more homophilic for $H(\mathcal{G})$ closer to $0$ or more heterophilic for  $H(\mathcal{G})$ closer to $1$.

Finally, we describe the semi-supervised node classification problem on event-based continuous graph as the following:
\begin{problem}\label{prob:tge}
   
  \textsc{Semi-supervised node classification on event-based continuous graph}.

  \textbf{Given:} (1) an event-based continuous graph $\mathcal{T}$, including \begin{itemize}
      \item 
  [(i)] a set of temporal events 
  $\{(v_i, e_{v_i, v_j}^t, v_j, t)\}$, where $v_i$ and $v_j$ are nodes, $e_{v_i, v_j}^t$ is the interaction (i.e., edge) between nodes and $t$ is the occurring moment of $e_{v_i, v_j}^t$; \item [(ii)] temporal node features: $\mathbf{x}_{v_i^t}$, $\mathbf{x}_{v_j^t}$, representing the extracted features of $v_i$ and $v_j$ at time $t$; \item [(iii)] \textbf{(Optional:)} temporal edge feature: $\mathbf{m}_{e_{v_i, v_j}^t}$, representing the extracted feature of $e_{v_i, v_j}^t$. 
  \end{itemize} (2) a training set of nodes with labels at specific time $t$: $\{y_{v_j}^{t}\}$.\\
  \textbf{Output:}  for any node $v$ at any time $t'$, the embedding $\mathbf{h}_{v^{t'}}$ of node $v$ at $t'$ and its predicted class $y_{v}^{t'}$.
\end{problem}

%% file: 05model.tex
\section{Model}\label{sec:model}
In this section, we present the \thgcn\ model. Firstly, we propose \textit{temporal edge heterophily measurement}, which measures the heterophily in event-based continuous graphs. Secondly, based on the definition of \textit{temporal edge heterophily measurement}, the key design of \thgcn\ is introduced: it adopts the low/high-pass signal filtering to handle the co-existence of the \textit{edge heterophily issue} and the \textit{temporal heterophily issue}. Thirdly, in-depth details of each module are presented. At last, we have a brief complexity analysis on \thgcn\ and a discussion on the connections between (1) \thgcn\ and static heterophilic graph neural networks, and (2) \thgcn\ and existing temporal graph neural networks for event-based continuous graphs.

\subsection{Temporal Edge Heterophily}\label{subsec:tgh}
To propose a model to handle complex event-based continuous graphs faced with the \textit{temporal edge heterophily challenge}, it is important to obtain a proper definition of \textit{temporal edge heterophily measurement}. Since edge heterophily and temporal heterophily co-exist in event-based continuous graphs, it is inadequate to directly clone the edge heterophily measurement from static graphs. The heterophily in both spatial (edge) and temporal domains should be considered simultaneously into such a measurement. Thus, we propose the \textit{temporal edge heterophily measurement} as follows:
\begin{equation}
    H(t_1, t_2, v_j) = \frac{\sum_{t\in[t_1, t_2)}{<y^{t_2}_{v_j} \neq y^{t}_{v_i}>}}{\sum_{t\in[t_1, t_2)}{|\{e^{t}_{v_i, v_j}\}|}},
\label{eq:teh}
\end{equation}
where $y^{t_2}_{v_j}$ is the label of $v_j$ at time $t_2$, $v_i$ is the node has directed edge connected to $v_j$, and $\{e^{t}_{v_i, v_j}\}$ is the set of events/interactions occurring between nodes $v_i$ and $v_j$ during time interval $[t_1, t_2)$. From Eq.~\eqref{eq:teh}, we can observe that the target node is $v_j$ at time $t_2$ and $H(t_1, t_2, v_j)$ measures the ratio of interactions/edges which link nodes with different labels during $[t_1,t_2)$. The relation between temporal and static edge heterophily measurements are: firstly, the time $t_1$ and $t_2$ are variables of $H(t_1, t_2, v_j)$, which indicates that its value is non-stationary in different time intervals, even for the same node; secondly, while the label of node $v_i$ is fixed in the \textit{static edge heterophily measurement} ($y_{v_i}$), it can change with the time in the \textit{temporal edge heterophily measurement} ($y^{t}_{v_i}$) in Eq. \eqref{eq:teh}; thirdly, the existing \textit{static edge heterophily measurement} can be recovered as a special case of the \textit{temporal edge heterophily measurement} if the node labels are fixed with time and the time interval is set as $[0,\infty)$. 

\subsection{Key Idea}\label{subsec:keyidea}
The key idea of \thgcn\ is utilizing the low/high-pass signal filtering to address the co-existence of the \textit{edge heterophily issue} and the \textit{temporal heterophily issue}. For one thing, the low/high-pass filtering is a widely used technique in static heterophilic graph neural networks \cite{bo2021beyond,yan2024trainable,du2022gbk}. Specifically, the low-pass filtering targets at making embeddings of connected nodes become similar (e.g., $\mathbf{I} + \mathbf{D}^{-\frac{1}{2}}\mathbf{A}\mathbf{D}^{-\frac{1}{2}}$ in GCN \cite{kipf2016semi}) to capture the homophilic information of edges in the graph, while the high-pass filtering aims to make embeddings of connected nodes more distinguishable  (e.g., $\mathbf{I} - \mathbf{D}^{-\frac{1}{2}}\mathbf{A}\mathbf{D}^{-\frac{1}{2}}$), which is able to reflect the heterophilic information. Through adaptively adjusting the weights of the low/high-pass filters, these static heterophilic graph neural networks can well solve the \textit{edge heterophily issue} (\textbf{SMP block}). For another thing, according to the definition of temporal edge heterophily measurement in Eq.~\eqref{eq:teh}, the proposed \thgcn\ should also take the \textit{temporal heterophily issue} into consideration and encode the temporal information in the module (\textbf{TMP block}).

\subsection{Algorithm Details}\label{subsec:thgcn}
In this subsection, we present the details of \thgcn, which is composed of two components: a sampler and an aggregator. Without loss of generality, 
we show the process of learning the embedding of the target node $v_j$ at time $t'$ as an example and the algorithm is shown in Algorithm~\ref{al:thgcn}.

\textbf{Sampler.} For node $v_j$ at time $t'$, we first set the time interval as $[t_0, t')$, where the choice of the starting point $t_0$ is a hyper-parameter. Then, we adopt the parallel sampler in TGL \cite{zhou2022tgl} ($\mathtt{Sampler}(v^{t'}_{j}, h_{\textrm{max}}, [t_0, t'), N_{\textrm{max}})$) to randomly sample a maximal $N_{\textrm{max}}$ number of interactions/edges $e^{t}_{v_i, v_j}$ within $h_{\textrm{max}}$ hops of $v_{j}$, where $t\in[t_0, t')$ \footnote{If the number of interactions/edges within $h_{\textrm{max}}$ hops of $v_{j}$ is smaller than $N_{\textrm{max}}$, all interactions/edges are sampled.}. All events/interactions involved in the described process form the interaction set $\mathcal{S}_{v_j}^{t'}=\{e^t_{v_i,v_j}\}$ for $v^{t'}_{j}$. For each interaction $e^t_{v_i,v_j}$ in $\mathcal{S}_{v_j}^{t'}$, we leverage the following time encoder to encode the time information, which has been validated to be effective in \cite{cong2022we}:
\begin{equation}
    E(t' - t) = \cos(\mathbf{w}(t' - t)),
\label{eq:te}
\end{equation}
where $\cos(\cdot)$ is the cosine encoding function and $\mathbf{w}$ is a trainable parameter vector. At last, $\mathcal{S}_{v_j}^{t'}$ contains the temporal information, node features, edge features, and the interaction/edge topology.
\begin{algorithm}[htbp]
\caption{\thgcn: Training Process for the Target Node $v_j^{t'}$.}
\label{al:thgcn}
\small
\KwIn{Event-based continuous graph $\mathcal{T}$;
\\
Temporal node features ${\mathbf{x}_{v_i^t}}$ for all nodes $v_i^t$ in $\mathcal{T}$;
\\
Node label of $v_j$ at $t'$: $y^{t'}_{v_j}$;
\\
Hyper-parameters: (1) time interval $[t_0, t')$; (2) maximum hops of interaction sampling $h_{\textrm{max}}$; (3) maximum numbers of sampled interactions $N_{\textrm{max}}$; and (4) layers $L$ of SMP block.
\\
\textit{Optional:} Temporal interaction/edge features: $\mathbf{m}_{e_{v_i, v_k}^t}$ for all interactions.}
\KwOut{The $L$-th SMP layer's embedding $\mathbf{h}^{(L+1)}_{v_j^{t'}}$}
Apply $\mathtt{Sampler}(v^{t'}_{j}, h_{\textrm{max}}, [t_0, t'), N_{\textrm{max}})$ to build $\mathcal{S}_{v_j}^{t'}$; \\
\For{each $e_{v_i,v_j}^t\in\mathcal{S}_{v_j}^{t'}$,$t\in[t_0, t')$}{
Encode time information with Eq.~\eqref{eq:te};\\
Produce the weight $p_{v^t_i,v^t_j}^{(0)}$ of the low-pass filter in TMP with Eq.~\eqref{eq:lowfilter};\\
Calculate $q_{v^t_i,v^t_j}^{(0)} = 1 - p_{v^t_i,v^t_j}^{(0)}$;\\
}
Calculate TMP output $\mathbf{h}^{(1)}_{v^{t'}_{j}}$ according to Eq.~\eqref{eq:firstlayer};\\
\For{$l \in \{1, 2, \dots, L\}$}{
\For{each $e_{v_i,v_j}^t\in\mathcal{S}_{v_j}^{t'}$, $t\in[t_0, t')$}{
Produce the weight $p_{v^t_i,v^t_j}^{(l+1)}$ of the low-pass filter in SMP with Eq.~\eqref{eq:secondlowfilter};\\
Calculate $q_{v^t_i,v^t_j}^{(l+1)} = 1 - p_{v^t_i,v^t_j}^{(l+1)}$;\\
}
Calculate the $l$-th layer of SMP output $\mathbf{h}_{v^{t'}_j}^{(l+1)}$ according to Eq.~\eqref{eq:finallayer};\\
}
Optimize $\mathbf{h}_{v^{t'}_j}^{(L+1)}$ via cross-entropy loss with $y^{t'}_{v_j}$.
\end{algorithm}

\textbf{Aggregator.}
With the sampled interaction set $\mathcal{S}_{v_j}^{t'}$, the aggregator of \thgcn\ conducts the message-passing operation to encode all related information. The aggregator is composed of two connected blocks: the temporal message-passing (TMP) block and the static message-passing (SMP) block \footnote{If the model does not have the SMP block, $l = 0$, which is equivalent to one-layer TMP block.}. The TMP block collects information from node features $\mathbf{x}_{v_i^t}$ and $\mathbf{x}_{v_j^t}$ of node $v_i$ and $v_j$ at $t$, the attribute $\mathbf{m}_{e_{v_i,v_j}^t}$ \footnote{The attribute of the interaction is optional.} of interaction $e_{v_i,v_j}^t$, and time encoding $E(t'-t)$. A multi-layer perceptron (MLP)~\cite{gardner1998artificial} is used to produce the weight $p_{v^t_i,v^t_j}^{(0)}$ of the low-pass filter for the temporal interaction $e_{v_i,v_j}^t$ happening to $v_j$ at $t'$:
\begin{equation}
    p_{v^t_i,v^t_j}^{(0)} = \sigma(\mathtt{MLP}([\mathbf{x}_{v_i^t}||\mathbf{x}_{v_j^t}||\mathbf{m}_{e_{v_i,v_j}^t}||E(t' - t)]),
    \label{eq:lowfilter}
\end{equation}
where $(\cdot||\cdot)$ is the concatenation and $\sigma(\cdot)$ is the sigmoid function to ensure $p_{v^t_i,v^t_j}^{(0)} \in (0, 1)$. In this way, the weight of the high-pass filter (i.e., $q_{v^t_i,v^t_j}^{(0)}$) can be obtained via the constraint: $ p_{v^t_i,v^t_j}^{(0)} +   q_{v^t_i,v^t_j}^{(0)} = 1$. Consequently, the output of the TMP block is calculated as follows:
\begin{equation}
    \mathbf{h}^{(1)}_{v^{t'}_j} = \mathbf{x}_{v_j^{t'}} + \frac{1}{|\mathcal{S}_{v_j}^{t'}|}\sum_{v_i^t}^{\mathcal{S}_{v_j}^{t'}}{(p_{v^t_i,v^t_j}^{(0)} -  q_{v^t_i,v^t_j}^{(0)})\mathbf{x}_{v_i^{t}}}
    \label{eq:firstlayer},
\end{equation}
where $ \mathbf{x}_{v_j^{t'}}$ is the node feature of $v_j$ at $t'$. 

Since the temporal information has already been encoded via the TMP block, we can then treat $\mathcal{S}_{v_j}^{t'}$ as a static graph \footnote{The same node at different moments are viewed as different nodes in this static graph.} with $\mathbf{h}^{(1)}_{v^{t'}_j}$ as the node feature \footnote{We can obtain $\mathbf{h}_{v^t_i}^{(1)}$s via similar TMP blocks for $v_i^t$s to $\mathbf{h}^{(1)}_{v^{t'}_j}$.}and add a subsequent SMP block to enhance the model's capability to process the spatial information. The weight of the low-pass filter is given by
\begin{equation}
     p_{v^t_i,v^t_j}^{(l+1)} = \sigma(\mathtt{MLP}([\mathbf{h}_{v^t_i}^{(l)}||\mathbf{h}_{v^t_j}^{(l)}]))\in (0, 1).
    \label{eq:secondlowfilter}
\end{equation}
This SMP block can contain multiple message-passing layers, in which the $l$-th layer's node embedding is calculated as
\begin{equation}
   \mathbf{h}_{v^{t'}_j}^{(l+1)} =   \mathbf{h}_{v^{t'}_j}^{(l)} + \frac{1}{|\mathcal{S}_{v_j}^{t'}|}\sum_{v^t_i}^{\mathcal{S}_{v_j}^{t'}}{(p_{v^t_i,v^t_j}^{(l)} - q_{v^t_i,v^t_j}^{(l)})\mathbf{h}_{v^t_i}^{(l)}}
   \label{eq:finallayer},
\end{equation}
where $\mathbf{h}_{v^t_i}^{(l)}$/$\mathbf{h}_{v^t_j}^{(l)}$ is the embedding of node $v_i$/$v_j$ at time $t$. The output node embedding of the SMP block is fed into a classifier. We optimize the \thgcn\ with cross-entropy loss in the semi-supervised temporal node classification task.
\subsection{Complexity Analysis \& Discussion.}
\textbf{Complexity analysis.} Here we give a brief time complexity analysis for the proposed \thgcn. The time complexity of the sampler in each epoch is $\mathcal{O}(|\mathcal{E}|)$ according to \cite{zhou2022tgl}, where $\mathcal{E}$ is the interaction set in the event-based continuous graph $\mathcal{T}$. For the aggregator, one message passing layer in either the TMP or the SMP block has time complexity $\mathcal{O}(|\mathcal{E}|\times f)$, where $f$ is the output dimension of the corresponding layer. 

\textbf{Discussion.} \thgcn\ can be viewed as a generalization of some existing, well-established static heterophilic graph neural networks and temporal graph neural networks. We illustrate this connection with some examples from these two categories. For static heterophilic graph neural networks, if we remove the temporal information and the interaction information from Eq.~\eqref{eq:lowfilter} and Eq.~\eqref{eq:firstlayer}, \thgcn\ will degrade to FAGCN~\cite{bo2021beyond} and the gating block of GBK-GNN~\cite{du2022gbk}. For temporal graph neural networks, if we exclude the process of learning the low-pass/high-pass weights and conduct the message-passing mechanism with equal weights, \thgcn\ will become the node encoder in GraphMixer~\cite{cong2022we}. If the low/high-pass weight in \thgcn\ is limited to single non-negative attention weight, \thgcn\ becomes similar to TGAT~\cite{xu2020inductive}. We provide ablation studies on the single non-negative attention weight in the experiment section.

%% file: 06experiment.tex
\section{Experiment}\label{secexp}
In this section, we evaluate the performance of \thgcn\ under the setting of \textit{semi-supervised temporal} node classification.

\subsection{Experiment Setup}
\textbf{Datasets.} We use 5 datasets for evaluation, including 3 traffic datasets: PEMSBA, PEMSLA and PEMSSD, where labels of nodes change with time, 1 social network dataset Reddit \cite{fan2021gcn} and 1 biological dataset Brain, where labels of nodes do not change with time. The detailed description of these datasets are attached in Appendix due to page limit. All the statistics are presented in Table~\ref{tab:dataset}. 
In addition, a metric named temporal changing ratio is proposed. This metric measures the ratio of nodes whose labels have changed. We can see that over 95\% of nodes have changed the labels during this period, which also matches the motivation of this work. The labels of nodes in Reddit and Brain are constructed by \cite{fan2021gcn} and remain the same during the whole time period, which is reflected in Table~\ref{tab:dataset} that the temporal changing ratios are 0s. We randomly split labeled nodes \footnote{In PEMSBA/PEMSLA/PEMSSD, since labels for the same node change with time, the same node with different labels at different times are viewed as two labeled nodes. In Reddit and Brain, the same node only appears once in the training/validation/test set.} in every dataset into 60/20/20\% for training, validation, and testing. 
\begin{table*}[htbp]
    \centering
    \caption{The statistics of datasets.}
    \begin{tabular}{c|ccccc}
        \hline
        Datasets & Reddit & Brain & PEMSBA & PEMSLA & PEMSSD\\
        \hline
        \#Nodes & 1,128  & 5000 & 1,631 & 2,377 & 673\\
        \#Events & 264,050 & 1895488 & 297,463 & 617,988 & 200,943\\
         Edge heterophily & 72.77\% & 78.91\% & 42.13\% & 34.61\% & 68.04\% \\ 
        \#Features of nodes & 20 &20 & 10 & 10 & 10 \\
        \#Classes & 4 & 10 & 5 & 5 & 5\\
        Temporal changing ratio & 0\% & 0\% & 95.70\% & 97.90\% & 97.33\% \\
        \hline
    \end{tabular}
    \label{tab:dataset}
\end{table*}


\textbf{Baselines and metrics.} We compare our model with 10 baseline methods, which can be divided into two groups: methods in the first group are temporal graph neural networks, including TGN \cite{rossi2020temporal}, JODIE \cite{kumar2019predicting}, APAN \cite{wang2021apan}, DySAT \cite{sankar2020dysat} and TGAT \cite{xu2020inductive}. Baselines in the second group are representative static heterophilic/homophilic graph neural networks: GCN \cite{kipf2016semi}, APPNP \cite{klicpera2018predict}, FAGCN \cite{bo2021beyond}, GPRGNN \cite{chien2020adaptive} and GAT \cite{velivckovic2017graph}. We adopt node classification accuracy as the metric for all methods.
\begin{table*}[t]
    \centering
    \caption{Performance comparison (mean$\pm$std accuracy (\%)) on five datasets.}
    \begin{tabular}{c|ccccccc}
        \hline
        Datasets & PEMSBA & PEMSLA & PEMSSD & Reddit & Brain\\
        \hline
        TGN  & 37.10$\pm 0.87$ & 48.60$\pm 0.66$ & 27.96$\pm 1.62$ & 15.78$\pm 1.05$ & 15.78 $\pm 1.05$\\
        JODIE  & 27.41$\pm 4.09$ & 37.03$\pm 5.53$ & 23.66$\pm 2.34$ & 30.80$\pm 1.99$ & 23.38 $\pm 3.44$\\
        APAN  & 38.04$\pm 1.07$ & 48.27$\pm 2.85$ & 29.97$\pm 1.01$ & 31.95$\pm 2.18$ & 27.88 $\pm 1.05$\\
        DySAT  & 31.93$\pm 1.13$ & 40.02$\pm 1.02$ & 30.73$\pm 0.86$ & 31.77$\pm 1.96$ & 16.24 $\pm 1.35$\\
        TGAT & 34.07$\pm 1.46$ & 45.39$\pm 0.70$ & 37.78$\pm 2.87$ & 29.29$\pm 4.72$ & 53.08 $\pm 2.01$\\
        APPNP & - & - & - & 30.44$\pm 1.78$ & 67.66 $\pm 1.22$\\
        GAT  &  - & - & - & 30.27$\pm 1.62$ & 56.66 $\pm 2.97$\\
        FAGCN  &  - & - & - & 30.18$\pm 2.35$ & 68.80 $\pm 1.19$ \\
        GCN  &  - & - & - & 29.73$\pm 3.08$ & 44.54 $\pm 1.49$ \\
        GPRGNN &  - & - & - & 31.24$\pm 1.87$ & 52.72 $\pm 1.25$\\
        \thgcn & \textbf{39.20$\pm \mathbf{1.23}$} & \textbf{55.82$\pm \mathbf{0.57}$} & \textbf{38.89 $\pm\mathbf{1.29}$} & \textbf{33.19$\pm \mathbf{4.09}$} & \textbf{69.18 $\pm \mathbf{1.76}$}\\
        \hline
    \end{tabular}
    \label{tab:performance}
\end{table*}

\textbf{Implementation details.} All hyper-parameters are set with a grid search. For the five datasets: PEMSBA, PEMSLA, PEMSSD, Reddit and Brain: (1) the layer of \thgcn\ is set as \{2, 2, 3, 1, 2\} respectively; (2) the number $N_{\textrm{max}}$ of sampled neighbors in one layer is set as \{10, 10, 5, 10, 40\}; (3) the number of epochs is set as \{100, 100, 150, 30, 100\}; and (4) the learning rate is set as \{0.01, 0.01, 0.01, 0.005, 0.01\}. All the results are the average over 5 runs and we report both the accuracy (ACC) and the standard deviation (std). All the experiments are run on a Tesla-V100 GPU.
\subsection{Results and Analysis}
The temporal node classification results are shown in Table~\ref{tab:performance} \footnote{Since labels of nodes change with time, static GNNs can not be applied to these three datasets, marked with `-' in the table.} . We have the following observations. Firstly, \thgcn\ outperforms all other 
temporal graph neural networks on various types of datasets, including both the spatio-temporal traffic data, PEMSBA/PEMSLA/PEMSSD, and the other two datasets, Reddit and Brain. This demonstrates the superiority of \thgcn\ to more effectively filter the homophilic/heterophilic graph signals under time-variant settings. Secondly, compared to the GNN models designed for static graphs (GCN/APPNP/GAT) or specifically tailored for heterophilic graphs (FAGCN/GPRGNN), which are incapable of dealing with the traffic data, \thgcn\ also shows clear advantages on Reddit and Brain. This reflects the effective utilization of temporal information by \thgcn\ over the static graph models. Overall, \thgcn\ is able to cover a variety of use cases, from homophilic graphs to heterophilic graphs and from datasets with time-invariant labels to datasets with time-variant labels, while gaining performance advantage in each case. 

\begin{figure}[t]
    \begin{center}
    \subfigure[Ablation study.]{\includegraphics[scale=0.2]{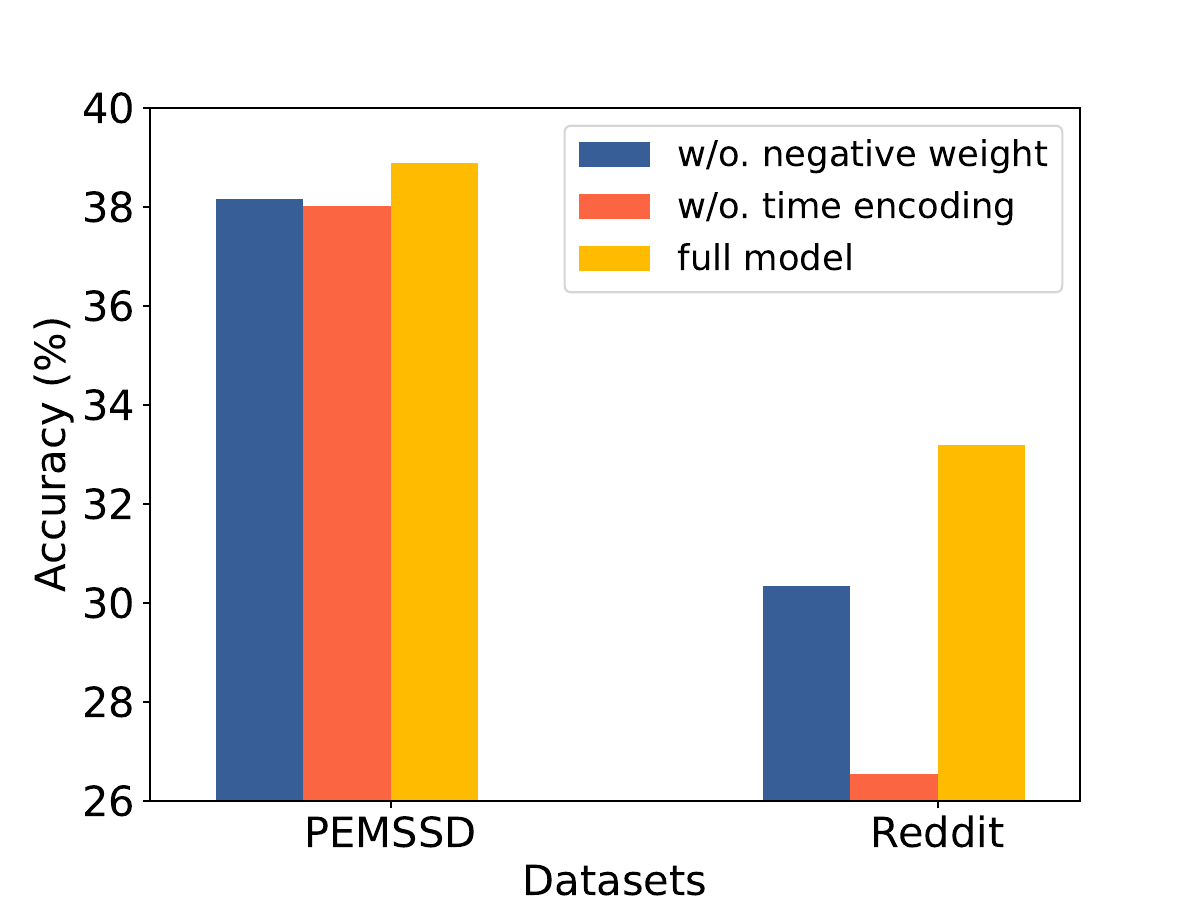}}
    \subfigure[Attention pair analysis.]{\includegraphics[scale=0.25]{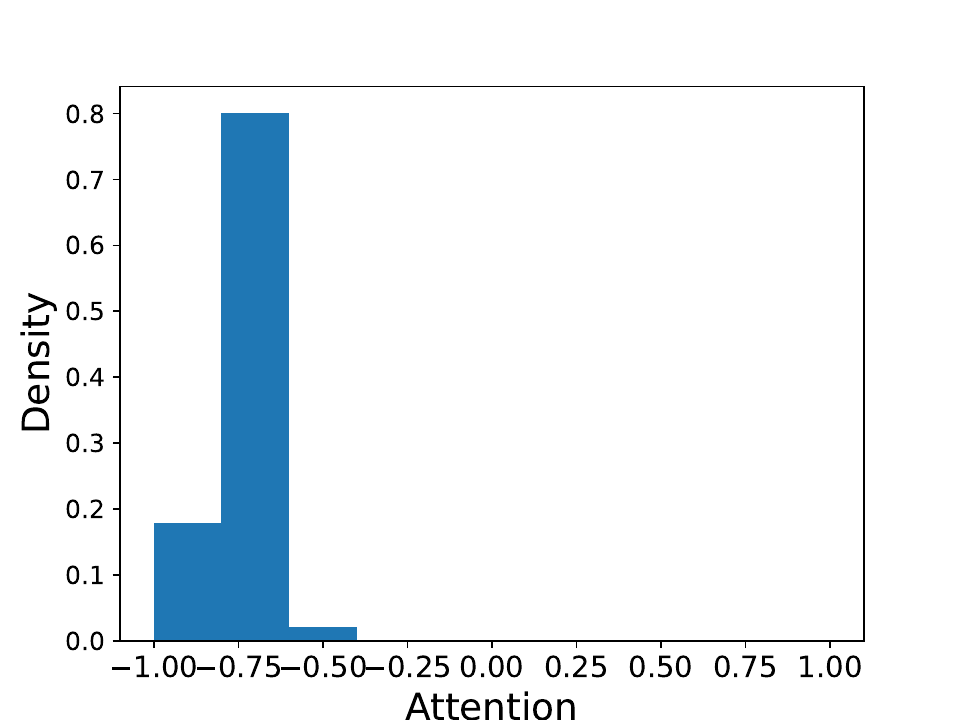}}
       \caption{Ablation study \& attention pair analysis.}
        \label{fig:as}
        \end{center}
\end{figure}

\textbf{Ablation study.} To better understand the reason behind the effectiveness of \thgcn,  we conduct an ablation study. First, \thgcn\ contains the low-pass and high-pass attention pair (i.e., $p_{v^t_i,v^t_j}^{(l)} - q_{v^t_i,v^t_j}^{(l)}$), which is negative if $p_{v^t_i,v^t_j}^{(l)} < q_{v_i^t,v_j^t}^{(l)}$. It is implemented via a $\mathtt{tanh}(\cdot)$ function. In the ablation study, we replace the $\mathtt{tanh}(\cdot)$ function with a $\mathtt{sigmoid}(\cdot)$ function to limit the attention to be a non-negative value, which corresponds to the \thgcn\ w/o negative weight in Figure~\ref{fig:as} (a). We can observe that the accuracy has dropped without negative weight. Specifically, for the PEMSSD dataset, \thgcn\ w/o. negative weight has similar accuracy (38.02\%) to the baseline TGAT (37.78\%). For the Reddit dataset, the performance of \thgcn\ without negative weight is 30.35\%, which is very close to that of GAT (30.27\%). The observed similarity in performances aligns with the fact that TGAT and GAT also deploy non-negative attention weights. Second, to demonstrate the effect of the time encoding in Eq.~\eqref{eq:te}, we remove it from \thgcn, which produces the variant of \thgcn\ w/o. time encoding. Figure~\ref{fig:as} (a) shows that time encoding is a component of great importance to \thgcn\, since it boosts its performance by 1.87\% in PEMSSD and remarkably by  6.64\% in Reddit. Therefore, the efficacy of the negative attention weight and the time encoding components shows they are indispensable to \thgcn. Third, we conduct an analysis on the attention pairs in the TMP block (i.e., $p^{(0)}_{v^t_i,v^t_j} - q^{(0)}_{v^{t}_i,v^{t}_j}$) to validate the temporal edge heterophily. For the PEMSBA dataset, the distribution of attention pairs for most recent interactions (i.e., $t' - t_0 < 5$) is shown in Figure~\ref{fig:as} (b). The x-axis is $p^{(0)}_{v^t_i,v^t_j} - q^{(0)}_{v^{t}_i,v^{t}_j}$ and the y-axis is the density/distribution of edges/interactions. We can observe that most attention pairs fall in the interval [-0.8, -0.6], which demonstrates that these most recent interactions actually have negative effects on the central node, which is consistent with the proposed temporal edge heterophily. 

\begin{figure}[t]
    \begin{center}
    \subfigure[\# of sampled neighbors.]{\includegraphics[scale=0.2]{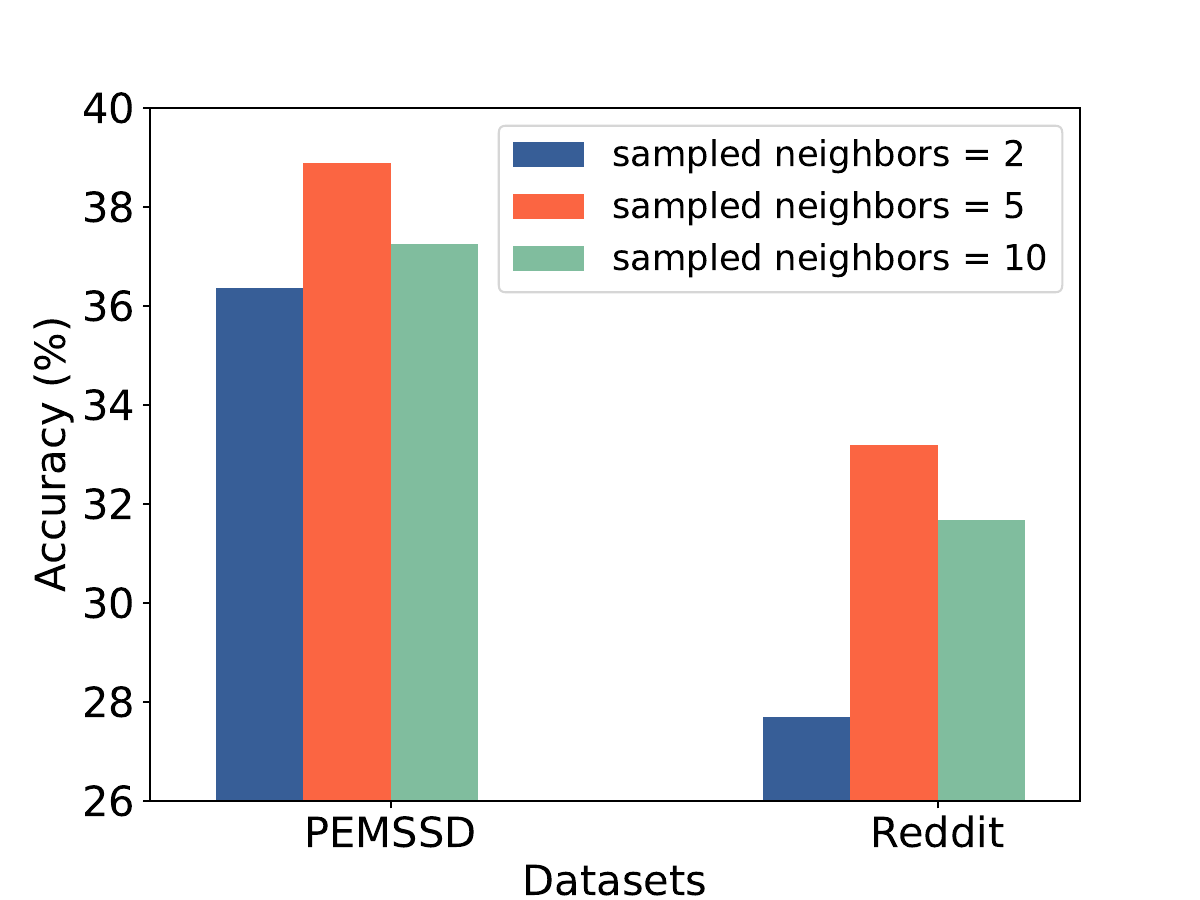}}
    \subfigure[Layers of \thgcn.]{\includegraphics[scale=0.2]{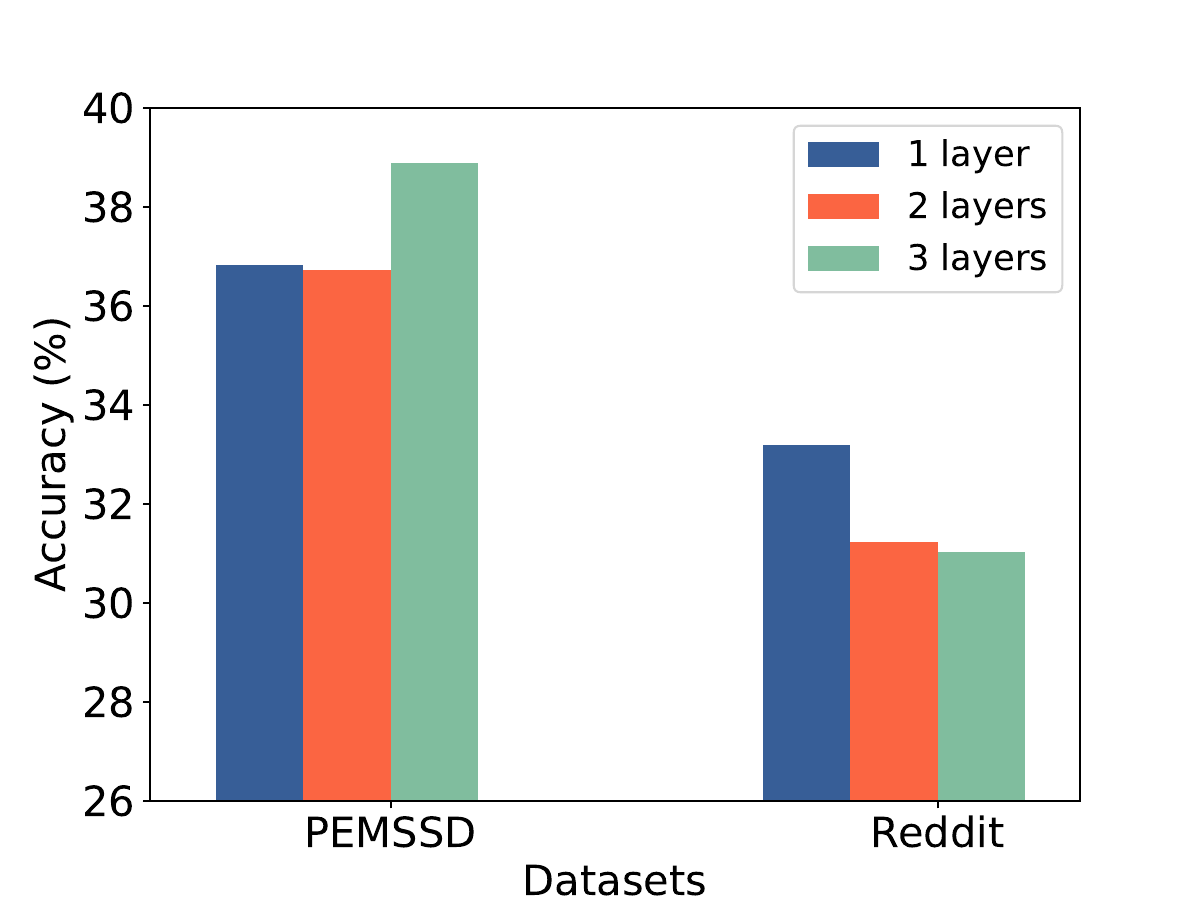}}
        \label{fig:study}
        \caption{Parameter study.}
        \label{fig:ps}
        \end{center}
    \vspace{-10pt}
\end{figure}

\textbf{Parameter study.} To better understand the performance sensitivity to the hyper-parameters, especially the number of sampled neighbors in each layer and the number of layers in \thgcn, systematic studies on these two hyper-parameters are performed. We set the number of sampled neighbors from \{2, 5, 10\} and set the layers of \thgcn\ from \{1, 2, 3\} on PEMSSD and Reddit. The results are presented in Figures~\ref{fig:ps} (a) and (b). From Figure~\ref{fig:ps} (a), we can observe that for PEMSSD, the best number of sampled neighbors in each layer is 5. Similarly, the best number of sampled neighbors in each layer for Reddit is also 5. This indicates that an appropriate reception field can well capture the information contained in interactions. On the other hand, for the hyper-parameter of layers, it varies across these two different datasets, which is shown in Figure~\ref{fig:ps} (b). Three layers are the best for the PEMSSD dataset, with 1 and 2 layers sharing similar performances, while for the Reddit dataset, the optimal option is only one layer. The accuracy decreases as the number of layers increases. This indicates that more layers involve higher-order neighbors in the reception field, which may act as noise once the layer number passes its optimal value.

%% file: 03related_work.tex
\vspace{-10pt}
\section{Related Works}
\textbf{Dynamic graph neural network (GNN).} According to the types of dynamic graphs, dynamic graph neural networks (GNNs) can be divided into discrete-time dynamic GNNs~\cite{goyal2018dyngem,wang2020streaming}, and event-based continuous GNNs (a.k.a. temporal GNNs)~\cite{nguyen2018continuous,rossi2020temporal,trivedi2019dyrep,cong2022we,zhou2022tgl}. Earlier dynamic graph neural networks \cite{goyal2018dyngem,wang2020streaming,you2022roland,bai2020adaptive} view the dynamic graph neural network as static graph snapshots, whose information is relatively coarse and thus is not the focus of our paper. Recently, a large number of temporal GNNs emerge to capture the information in event-based continuous graphs. To name a few, to capture the temporal dynamics, temporal point process is formulated in DyRep \cite{trivedi2019dyrep}. Temporal motifs are adopted in CAW \cite{wang2020streaming} to model the dynamics of the graph. TGAT~\cite{xu2020inductive} transfers the attention mechanism from GAT~\cite{velivckovic2017graph} in static graphs to event-based continuous graphs. TGN~\cite{rossi2020temporal} is designed to memorize long term dependencies efficiently and TGL~\cite{zhou2022tgl} optimizes the sampling process of existing event-based continuous graphs. GraphMixer~\cite{cong2022we} simplifies the complex structures of existing event-based continuous GNNs. However, all of the above event-based continuous GNNs are developed based on the temporal homophily assumption. The latest work GRETO ~\cite{zhou2022greto} notices the temporal heterophily issue, but it is a discrete-time dynamic GNN, which can not be applied to event-based continuous graphs. To our best knowledge, \thgcn\ is the first work on temporal heterophily in event-based continuous graphs.


\textbf{Node classification on static heterophilic graphs.} Most existing graph neural networks (GNNs)~\cite{kipf2016semi,velivckovic2017graph,hamilton2017inductive} are developed based on the homophily assumption, referring to the phenomenon that connected node pairs tend to share the same labels. Recently, researchers pay more attention to heterophilic graphs, where connected node pairs tend to have disparate labels. From the spatial aspect, H2GCN \cite{zhu2020beyond} and Geom-GCN \cite{pei2020geom} adopt a large neighborhood field for message-passing. From the spectral aspect, FAGCN \cite{bo2021beyond} and GPRGNN \cite{chien2020adaptive} adaptively integrate high/low frequency signals with trainable parameters. 
In addition, HOG-GCN \cite{wang2022powerful} and CPGNN \cite{zhu2021graph} propose an alternative message-passing mechanism to handle the homophily and the heterophily. Recent progresses in this domain include ACM-GCN \cite{luan2021heterophily,luan2022revisiting}, LINKX \cite{lim2021large}, BernNet \cite{he2021bernnet}, GloGNN \cite{li2022finding}, and GBKGNN \cite{du2022gbk}. \cite{zheng2022graph} has conducted a comprehensive survey on this topic.

%% file: 07conclusion.tex
\section{Conclusion and Limitation}
This paper tackles the limitations observed in existing temporal graph neural networks through the introduction of the innovative Temporal Heterophilic Graph Convolutional Network (\thgcn). Firstly, we propose a novel temporal edge heterophily measurement, extending the conventional static edge heterophily metrics. Secondly, departing from conventional methods grounded in the homophily assumption, our model adeptly captures the nuanced dynamics of real-world event-based continuous graphs by addressing both edge heterophily and temporal heterophily. The \thgcn\ model incorporates the high/low-pass graph signal filtering technique from static heterophilic graphs and comprises two integral components: (1) a sampler and (2) an aggregator. Thirdly, extensive experiments demonstrate its superb performance in semi-supervised temporal node classification tasks across five diverse real-world datasets. The results underscore the efficacy and versatility of the proposed \thgcn\ model in navigating the complexities of event-based continuous graphs. For the limitation of our work, since we only focus on the node classification task, other tasks (e.g., the temporal link prediction task) need to be further explored.

%% file: 08appendix.tex
\section{Appendix}
\subsection{Construction of Datasets}
\label{subsec:con_dataset}
For $3$ traffic datasets, the raw data is collected from the Caltrans Performance Measurement System (PeMS) \cite{chen2001freeway}, which is composed of the freeway system in major areas of California. The traffic statistics from three areas are chosen: Bay Area (Station 4), Los Angeles (Station 7), and San Diego (Station 11) to build the corresponding PEMSBA, PEMSLA, and PEMSSD datasets in our paper. In detail, for each dataset, each sensor is considered as a node. A 120-minute interval traffic flow data from each sensor ($v_i$ at time $t$) is collected. The total number of intervals is 84 (i.e., one week). Each interval contains three aspects of information: flow rate, speed, and occupancy of the sensor. We also have the location (i.e., latitude and longitude) of each sensor. For each $v_i$ at time $t$, we first select 5 nodes with the most similar flow rate. Then, a threshold is set (i.e., $\tau=0.000001$) to filter out sensors/nodes with Euclidean distance to $v_i$ larger than the threshold. The remaining sensors $v_j$s within the distance threshold form the interactions with $v_i$ at $t$. After constructing the interactions, we divide the range of speed values into 10 intervals with equal length. Depending on the interval of the speed value of $v_i$ at $t$, a 10 dimension one hot vector is built as the attribute $\mathbf{x}^{t}_{v_i}$, where only one entry is 1 and the remaining entries are 0. At last, we utilize the occupancy to build the labels of nodes. For all nodes at different moments, we collect the occupancy values of them during the last 12 intervals (i.e., the last day). Similar to the process of constructing node attributes. The occupancy values are divided into 5 intervals with the same length and each interval corresponds to one class. For these labeled nodes, the edge/interaction heterophily ranges from 34.61\% of PEMSLA to 68.04\% of PEMSSD. In addition, a metric named temporal changing ratio is proposed. This metric measures the ratio of nodes whose labels have changed during the last 12 intervals. We can see that over 95\% of nodes have changed the labels during this period, which also matches the motivation of this work. 

The Reddit dataset is generated from Reddit. The raw data of Reddit contains web content ratings, discussions among users, and some social news. To construct the Reddit dataset in our experiments,  The graph is constructed by viewing posts as nodes. If two posts at the same time intervals share the same keywords, we build an interaction between them. We apply word2vec \cite{mikolov2013distributed} on the post comments to generate the node attributes. The labels of nodes are constructed by \cite{fan2021gcn} and remain the same during the whole time period. The Brain dataset is generated from functional magnetic resonance imaging (fMRI) data \footnote{\url{https://tinyurl.com/y4hhw8ro}}. Nodes represent cubes of brain tissue, and two nodes are connected during a time period if they show similar degrees of activation during that time period.
We randomly split labeled nodes \footnote{In PEMSBA/PEMSLA/PEMSSD, since labels for the same node change with time, the same node with different labels at different times are viewed as two labeled nodes. In Reddit, the same node only appears once in the training/validation/test set.} in every dataset into 60/20/20\% for training, validation, and testing.